\documentclass{article}
\usepackage[utf8]{inputenc}

\title{A Retail Product Categorisation Dataset}
\author{Febin Sebastian Elayanithottathil and Janis Keuper }
\date{Institute for Machine Learning and Analytics\\
Offenburg University, Germany}

\usepackage{graphicx}
\usepackage{hyperref}
\usepackage{array}
\hypersetup{
    colorlinks=true,
    linkcolor=blue,
    filecolor=magenta,      
    urlcolor=cyan,
}

\begin{document}

\maketitle

\section{Introduction}
Most eCommerce applications, like web-shops have millions of products. In this context, the identification of similar products is a common sub-task, which can be utilized in the implementation of recommendation systems, product search engines and internal supply logistics. Providing this data set, our goal is to boost the evaluation of machine learning methods for the prediction of the category of the retail products from tuples of images and descriptions. 

\subsection{Data Set Description}
The retail products data set consist of around 48000 products from 21 categories with colour images (100x100) and the according description text. The data set is divided into 42000 training samples and around 6000 plus test samples.  The samples are equally distributed among 21 categories. The train and test folder of this dataset contains the product images saved by their unique id (ImgId). The train.csv file contains the following information.
\begin{itemize}
  \item ImgId: Unique ID of the product, e.g. 9966645691
  \item title: Name of the product
  \item description: Description of the product
  \item categories: Name of the category the product belongs to
\end{itemize}
All the products in this dataset are belonging to following 21 categories:\newline \newline
\textit{Electronics, Sports \& Outdoors, Cell Phones \& Accessories, Automotive, Toys \& Games, Tools \& Home Improvement, Health \& Personal Care, Beauty, CDs \& Vinyl, Grocery \& Gourmet Food, Office Products, Arts, Crafts \& Sewing, Pet Supplies, Patio, Lawn \& Garden, Clothing, Shoes \& Jewelry, Movies \& TV, Baby, Musical Instruments, Industrial \& Scientific, Baby Products, Appliances, All Beauty, All Electronics}\newline \newline

The dataset is created from a larger set provided by amazon review data (2018)\cite{dataset}. It is a large public dataset that includes reviews (ratings, text, helpfulness votes), product metadata (descriptions, category information, price, brand, and image features). \cite{jianmo}.

\begin{figure}[h!]
\centering
\includegraphics[width=.95\textwidth]{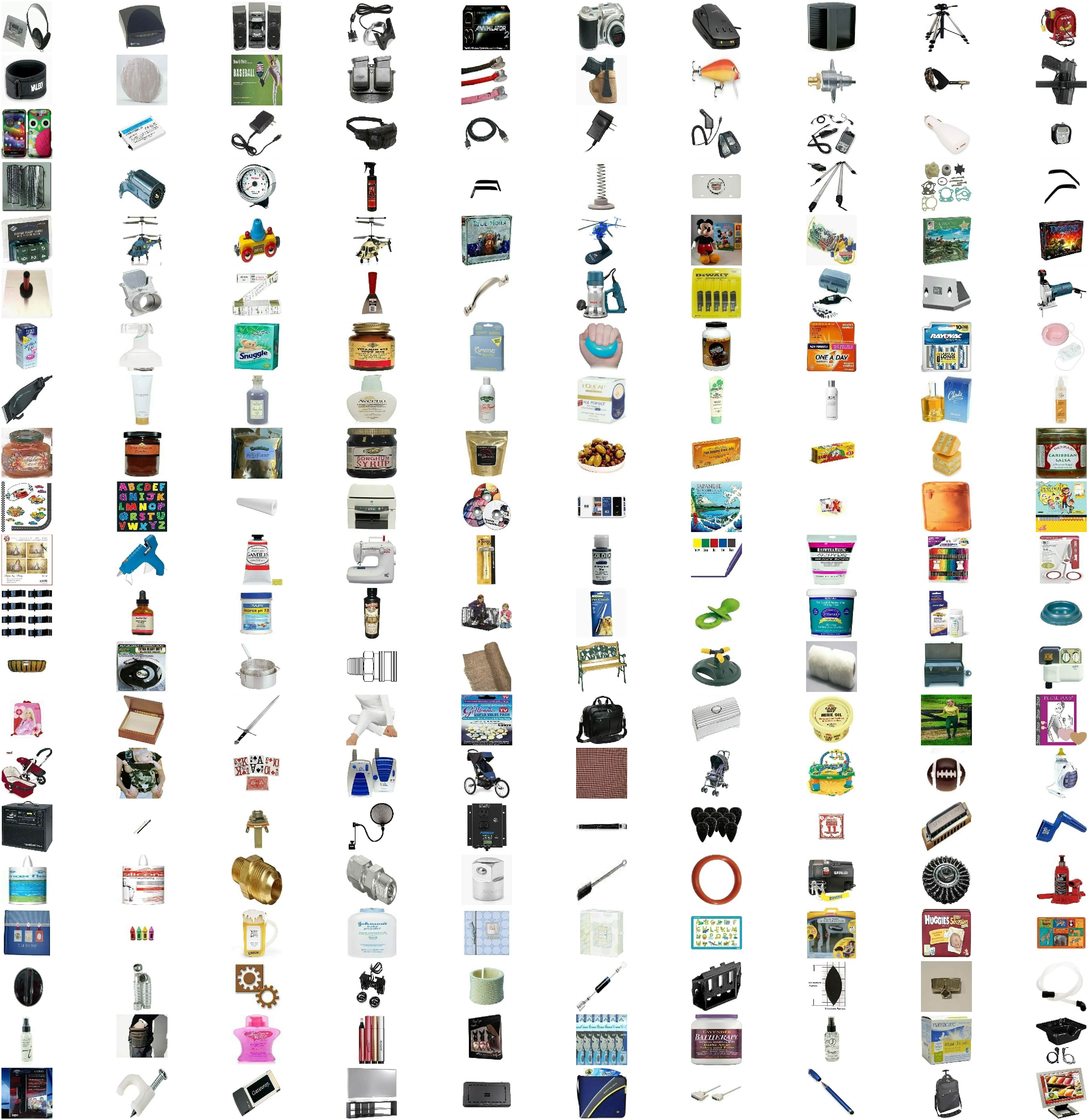}
\caption{Sample product images from all categories}
\label{fig:universe}
\end{figure}

\begin{table}[h!]
  \centering
  \begin{tabular}{ | c | m{7cm} | m{2cm} | }
    \hline
    Image & Description & Category \\ 
    \hline
    \begin{minipage}{0.17\textwidth}
      \includegraphics[width=20mm, height=20mm]{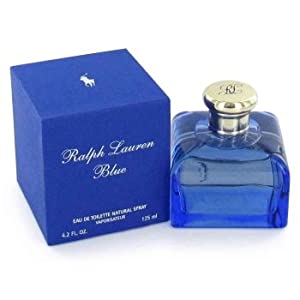}
    \end{minipage}
    &
    \parbox{7cm}{Authentic ANAIS ANAIS by Cacharel Perfume for Women. Manufactured by the design house of Cacharel.} & Beauty \\
    \hline
        \begin{minipage}{0.17\textwidth}
      \includegraphics[width=20mm, height=20mm]{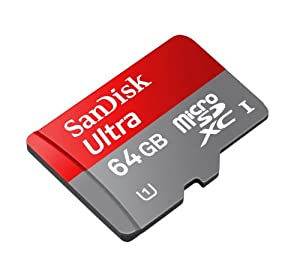}
    \end{minipage}
    &
    \parbox{7cm}{Professional Ultra SanDisk 64GB MicroSDXC Asus Transformer Book T100TA card is custom formatted for high speed, lossless recording! Includes Standard SD Adapter.}
    & 
    Electronics \\
    
    \hline
        \begin{minipage}{.17\textwidth}
      \includegraphics[width=20mm, height=20mm]{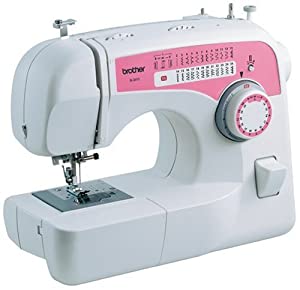}
    \end{minipage}
    &
    \parbox{7cm}{Brother's XL2610 is a 59 stitch function free arm sewing machine.  The free arm feature allows you to sew cuffs or pant legs easily and convert the machine back to a flat bed sewing area for larger items.  The machine features an automatic needle threader and built-in thread cutter to make setting up your machine easy.} & Arts, Crafts \& Sewing \\

 \hline
    \begin{minipage}{.17\textwidth}
      \includegraphics[width=20mm, height=20mm]{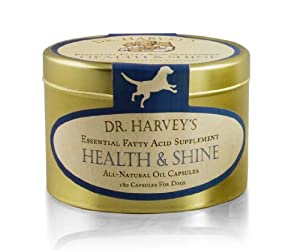}
    \end{minipage}
    &
    \parbox{7cm}{Health and Shine is pure salmon oil capsules. It is used as a supplement for skin and coat enhancement. In addition Health and Shine may help with heart health and circulation as well as joint stiffness and other inflammation in the body. Health and Shine provides Omega 3 fatty acids. Fatty acids have been found to be important for overall good health and well-being in dogs.} & Pet Supplies \\
  \hline
        \begin{minipage}{.17\textwidth}
      \includegraphics[width=20mm, height=20mm]{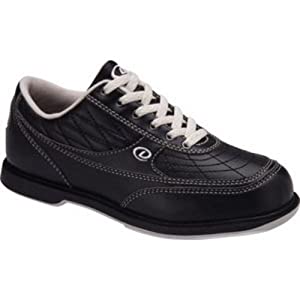}
    \end{minipage}
    &
    \parbox{7cm}{Soft Durable Man-Made Upper.Fully Fabric-lined with Padded Tongue and Collar.Non-Marking Classic Rubber Outsole Raised Heel Microfiber S8 Slide Soles on both shoes.} & Clothing, Shoes \& Jewelry \\
    \hline
  \end{tabular}
  \caption{Sample descriptions from dataset with their image and category }\label{tbl:Samples}
\end{table}

\subsection{Kaggle Challenge}
We provide the dataset via Kaggle\cite{kaggle}, alongside a public product categorization challenge. You can download this dataset from the kaggle challenge\cite{kaggleDataset}. The goal of the competition is to build a machine learning model to predict the category of retail products from their image and description.

\section{Baseline Solution}
Our baseline solution results on the data set are illustrated in the table \ref{tab:my_label}. These results are obtained by building a deep learning model by concatenating a Convolutional Neural Network (ConvNet) and a Long Short- Term Networks (LSTM) \cite{lstm}. ConvNet will classify the product images and the LSTM network with an embedding layer will classify the description text. We have used a supervised learning process to train the model by labelling the samples by their category\cite{thesis}. The neural network models learn while training by a feedback process called backpropagation. This involves comparing the output produced by the network with the actual output and using the difference between them to modify the weights of the connections between the units in the neural network. Our combined text and image classification model achieved around 71.81 F1 score on the test data. Figure 2 illustrates the embedding space in a two- dimensional space after applying the t-SNE algorithm. From this, we can see that most of the samples from similar category have similar vector representations. We provide a notebook\cite{notebook} in the kaggle challenge that helps to find our baseline results on this data.

\begin{table}[]
    \centering
    \begin{tabular}{ |c|c| } 
         \hline
         Accuracy & 74.57  \\ 
         \hline
         Validation Accuracy & 72.85  \\ 
         \hline
         Precision & 88.54  \\ 
         \hline
         Recall & 75.60  \\ 
         \hline
         F1 Score (Validation Data) & 81.56  \\ 
         \hline
         F1 Score (Test Data) & 71.56  \\ 
         \hline
    \end{tabular}
    \caption{Baseline results}
    \label{tab:my_label}
\end{table}

\begin{figure}[h!]
\centering
\includegraphics[width=0.95\textwidth]{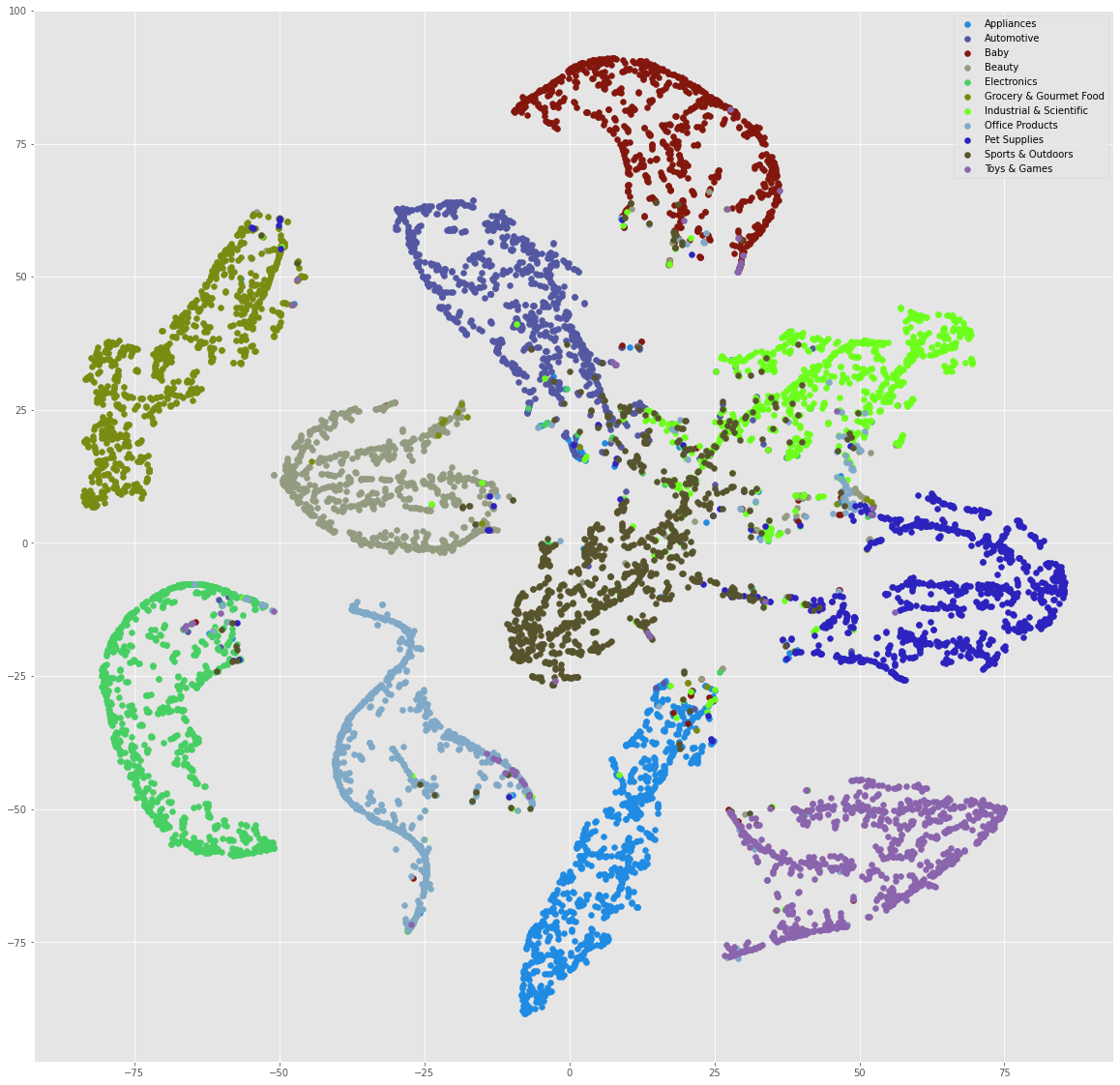}
\caption{t-SNE visualization of the embedding}
\label{fig:universe}
\end{figure}

\bibliographystyle{plain}
\bibliography{references}

\begin{thebibliography}{1}

\bibitem{notebook}
Baseline solution notebook.
\newblock
  \url{https://www.kaggle.com/imlahsoffenburg/retail-product-classification}.
\newblock Accessed: 2021-03-23.

\bibitem{kaggle}
Retail products classification.
\newblock \url{https://www.kaggle.com/c/retail-products-classification}.
\newblock Accessed: 2021-03-23.

\bibitem{kaggleDataset}
Retail products classification dataset.
\newblock \url{https://www.kaggle.com/c/retail-products-classification/data}.
\newblock Accessed: 2021-03-23.

\bibitem{thesis}
Febin~Sebastian Elayanithottathil.
\newblock Learning deep embedding for product representation (master's thesis).
\newblock 10 2020.

\bibitem{lstm}
Sepp Hochreiter and Jürgen Schmidhuber.
\newblock Long short-term memory.
\newblock {\em Neural computation}, 9:1735--80, 12 1997.

\bibitem{jianmo}
Julian~McAuley Jianmo~Ni, Jiacheng~Li.
\newblock Justifying recommendations using distantly-labeled reviews and
  fined-grained aspects.
\newblock {\em Empirical Methods in Natural Language Processing (EMNLP)}, 2019.

\bibitem{dataset}
Jianmo~Ni UCSD.
\newblock Amazon review data (2018).
\newblock \url{https://nijianmo.github.io/amazon/index.html}.
\newblock Accessed: 2021-03-23.

\end{thebibliography}
\end{document}